\documentclass[letterpaper, 10 pt, conference]{ieeeconf}  % Comment this line out if you need a4paper

\IEEEoverridecommandlockouts                              % This command is only needed if 
\usepackage{cite}
\usepackage{graphics} % for pdf, bitmapped graphics files
\usepackage{epsfig} % for postscript graphics files
\usepackage{amsmath} % assumes amsmath package installed
\usepackage{amssymb}  % assumes amsmath package installed
\usepackage{booktabs}
\usepackage{multirow} 
\usepackage{subcaption} % For subfigure environment
\usepackage{float}
\usepackage{placeins}
\usepackage{hyperref}
\usepackage{siunitx}
\usepackage[dvipsnames]{xcolor}

\usepackage[square, numbers, sort&compress]{natbib}

\title{\LARGE \bf
WildFusion: Multimodal Implicit 3D Reconstructions in the Wild
}

\author{Yanbaihui Liu$^{1}$ and Boyuan Chen$^{1}$ \\
\textcolor{orange}{\href{http://generalroboticslab.com/WildFusion}{www.generalroboticslab.com/WildFusion}} %<-this % stops a space
\vspace{-0.4cm}
\thanks{*This work is supported by DARPA TIAMAT HR00112490419, DARPA FoundSci HR00112490372, ARL STRONG W911NF2320182 and W911NF2220113. $^{1}$ All authors are from Duke University.}%
}

\begin{document}

\maketitle
\thispagestyle{empty}
\pagestyle{empty}

\setlength{\belowdisplayskip}{2pt}
\setlength{\textfloatsep}{4pt}	

%%%%%%%%%%%%%%%%%%%%%%%%%%%%%%%%%%%%%%%%%%%%%%%%%%%%%%%%%%%%%%%%%%%%%%%%%%%%%%%%
\begin{abstract}

We propose WildFusion, a novel approach for 3D scene reconstruction in unstructured, in-the-wild environments using multimodal implicit neural representations. WildFusion integrates signals from LiDAR, RGB camera, contact microphones, tactile sensors, and IMU. This multimodal fusion generates comprehensive, continuous environmental representations, including pixel-level geometry, color, semantics, and traversability. Through real-world experiments on legged robot navigation in challenging forest environments, WildFusion demonstrates improved route selection by accurately predicting traversability. Our results highlight its potential to advance robotic navigation and 3D mapping in complex outdoor terrains.

\end{abstract}

%%%%%%%%%%%%%%%%%%%%%%%%%%%%%%%%%%%%%%%%%%%%%%%%%%%%%%%%%%%%%%%%%%%%%%%%%%%%%%%%
\section{Introduction}

Robots need effective environmental representations to navigate safely and accomplish tasks successfully in unstructured outdoor environments -- often referred to as ``in-the-wild'' settings such as monitoring high-voltage power lines and extinguishing forest fires. However, accurately modeling these environments presents significant challenges due to their inherent complexity. The lack of clear boundaries between objects, combined with fluctuating lighting conditions and shadows, creates a dynamic and often ambiguous visual landscape. Additionally, these environments don't follow predefined patterns or rules for interpretation, hence creating models that can reliably understand and represent them becomes difficult.

Traditional 3D reconstruction methods, such as LOAM \cite{Zhang-2017-110808} and LIO-SAM \cite{shan2020lio}, typically rely on a single sensor modality like LiDAR or cameras. These methods have shown success in mapping static scenes. Other approaches \cite{bescos2018dynaslam, yu2018ds, chen2019suma++} employ semantic segmentation to detect and handle moving objects for improved performance in dynamic environments. However, both approaches struggle to provide high-quality and drift-free reconstruction in complex in-the-wild settings, where a single sensor modality is not enough to overcome the scene complexities and the inherent limitations of vision-based sensors.

Recent research has focused on addressing this issue by studying multimodal sensor fusion. This includes combining LiDAR and cameras \cite{lin2022r, kong2024multi}, or integrating additional modalities such as audio, language and, tactile data \cite{yue2020day, jatavallabhula2023conceptfusion, weerakoon2023graspe}. While these multi-sensing systems allow for scene reconstructions for more advanced tasks in complex environments, they still rely on explicit representations, such as point clouds \cite{wang2017sigvox}, voxels \cite{oleynikova2017voxblox}, or meshes \cite{zhang2021sketch2model}. Techniques such as Gaussian Splatting \cite{kerbl20233d, ji2024neds} have further advanced explicit methods for processing dynamic outdoor scenes with stronger robustness to noise and more effective handling sparse data. However, as other explicit representations, they still require dense data to produce accurate models.

\begin{figure}[t]
    \centering
    \includegraphics[width=0.75\linewidth]{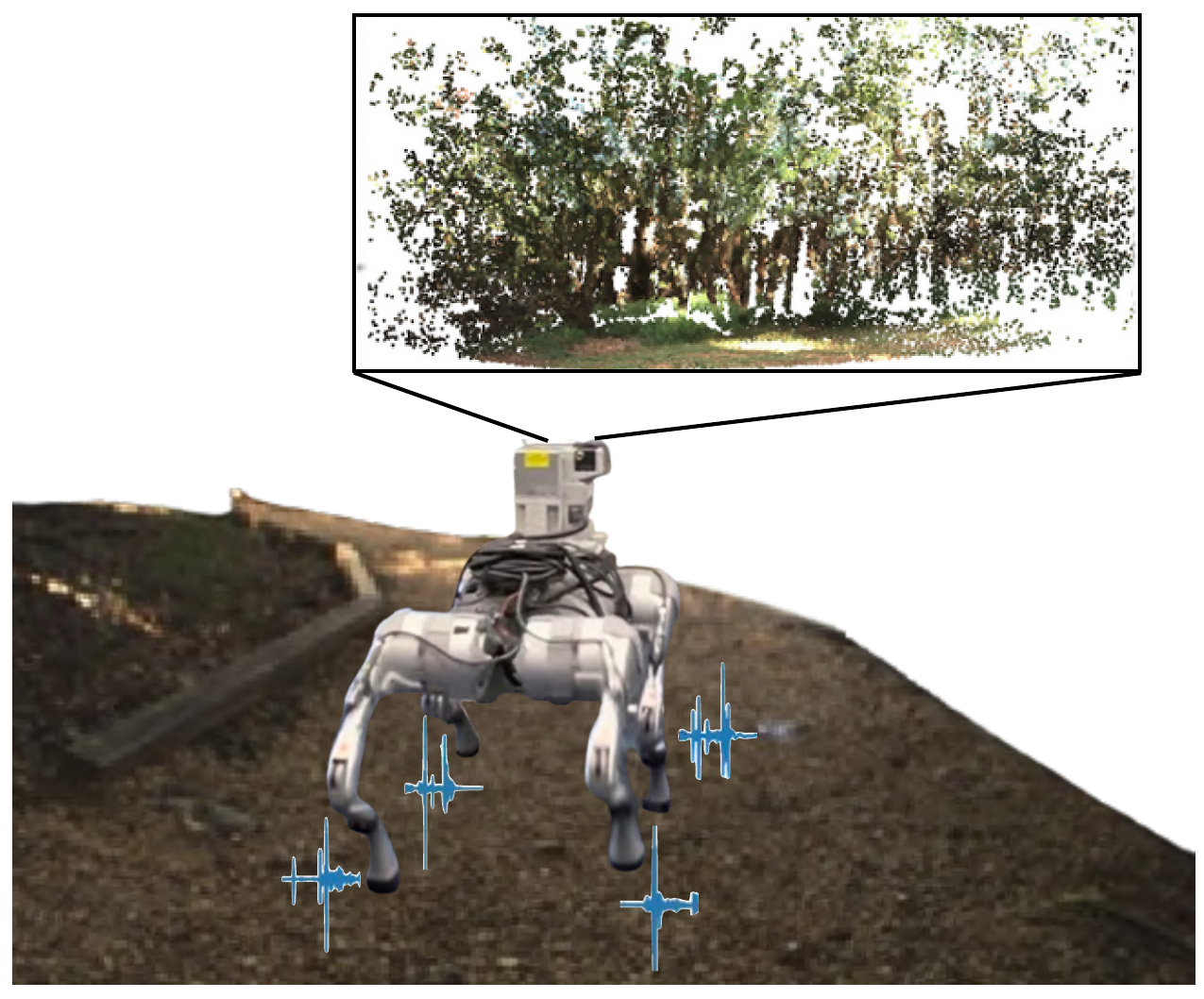}
    \caption{\textbf{WildFusion} integrates LiDAR, camera, microphones, and tactile sensors with implicit neural representations for continuous 3D scene reconstruction. The learned model provides accurate and continuous traversability predictions to enhance legged robot navigation in forest environments.}
    \label{fig:teaser}
\end{figure}

In recent years, implicit representation methods using Neural Radiance Fields (NeRF) \cite{mildenhall2020nerf, yu2021pixelnerf, barron2021mip, adamkiewicz2022vision} and Signed Distance Fields (SDFs) \cite{park2019deepsdf, cheng2023sdfusion} have received increasing attention. These approaches are particularly useful for describing complex surfaces and are more robust to sparse or incomplete data. Although existing implicit methods have made breakthroughs in 3D modeling and mapping, most solutions still rely heavily on visual or geometric sensor data. Research on fusing multimodal signals, such as acoustic and tactile data, remains limited, even though these modalities are highly useful for providing richer sensory information in unstructured environments.

We introduce WildFusion (Fig.~\ref{fig:teaser}), a multimodal implicit 3D reconstruction framework that integrates LiDAR, RGB camera, audio, and tactile sensors on a quadruped robot platform. WildFusion generates pixel-level continuous scene representations of multiple environmental features, including 3D geometry, semantics, continuous traversability, color, and confidence scores. By leveraging implicit representations, our method can produce complete scene representations from sparse inputs. We present a multimodal in-the-wild dataset by navigating our quadruped robot in a forest environment. Our experiments suggest that our multimodal formulation effectively leverages all modalities to enhance the 3D scene reconstruction with a richer understanding of the surrounding environment. By testing WildFusion on a downstream legged robot navigation task, we show that the robot can safely traverse various terrains such as grasslands, dense leaves, high vegetation, and gravel. Furthermore, since our model design is modular, we believe that WildFusion provides a promising direction to integrate more diverse sensing modalities to build rich and robust scene representations.

%%%%%%%%%%%%%%%%%%%%%%%%%%%%%%%%%%%%%%%%%%%%%%%%%%%%%%%%%%%%%%%%%%%%%%%%%%%%%%%%
\section{Related Work}

\subsection{3D Mapping}

Simultaneous Localization and Mapping (SLAM) is a key technique for mapping and navigating in unknown environments. SLAM enables robots to build maps of the environment while determining their positions. Traditional SLAM algorithms, such as direct methods \cite{engel2014lsd} and feature-based methods \cite{Mur2017ORB, Campos2021ORB}, have been successfully applied to city-view mapping. However, these methods often rely on salient environmental features which may not be available in unstructured environments. Moreover, they also produce sparse maps without sufficient details for more complex environments and tasks. Other works focus on generating dense maps \cite{newcombe2011kinectfusion, dai2017bundlefusion} to incorporate scene geometry and color to provide richer environmental information, but still leave gap and blank areas. Moreover, the use of explicit representations makes the system sensitive to outliers.

To address the above challenges, researchers have begun exploring implicit representations to enhance SLAM. Some approaches use radiance-based or coordinate-based representations \cite{sucar2021imap, zhu2022nice, wang2023co, liu2023robust} to effectively fill surface holes and represent both geometry and appearance such as color and texture in a photorealistic way. Others rely on SDF-based methods \cite{camps2022learning, sun2021neuralrecon, ortiz2022isdf, chen2022fully} for fine-grained surface reconstruction. Our approach builds on these advancements by leveraging implicit representations in SLAM to achieve continuous environmental mapping. However, we go further by incorporating multimodal signals beyond just vision with real-world legged robot demonstrations.

\subsection{Traversability Prediction}

Traversability prediction in outdoor settings has been addressed through several methods. The most straightforward and efficient approaches directly derive traversability from geometric representations or elevation maps \cite{wermelinger2016navigation, fan2021step, cao2022autonomous}. However, these techniques often struggle when faced with similar terrains in the same area. Additionally, a heavy reliance on geometry ignores textures and semantics, which leads to mistakes hindering smooth navigation in outdoor environments, such as classifying high vegetation as solid obstacles. Semantic-aided methods help mitigate these issues by assigning different traversability costs to distinct terrain classes \cite{leung2022hybrid, pmlr-v164-shaban22a, roth2023viplanner}. However, the predicted traversability values cannot handle fine-grained classes such as wet soil or dry land -- two surfaces that belong to the same high-level terrain classes but offer very different levels of ease for traversal.

More advanced methods predict traversability by considering the robot’s physical interactions with the environment. For example, contrastive learning with human driving data and segmentation masks \cite{jung2023v} has been used to predict traversability in both on- and off-trail settings. Terrain semantics can also be mapped to vehicle speed profiles \cite{cai2022risk}, allowing robots to adapt their speed in real time as terrain conditions change. Other approaches integrate vision and confidence estimation to account for deviations between expected and actual speeds, improving navigation in uncertain environments \cite{mattamala2024wild}. Additionally, risk-aware frameworks penalize uncertain terrains using evidential learning to enhance decision-making by factoring in traction predictions \cite{cai2024evora}. Our method also leverages the physical interactions with the environment from a legged robot, but instead uses proprioception signals to compute ground-truth continuous traversability scores. Our design enables us to generate pixel-level traversability predictions that accurately reflect the robot's interaction with the environment.

\subsection{Acoustic Profiling}

Sound has emerged as a valuable modality for object perception. Previous research has shown that feedback from audio synthesis engines can be used to predict physical parameters of objects \cite{zhang2017shape}, and acoustic vibrations can be used to sense real-world object properties \cite{chen2021boombox, liu2024sonicsense}. In terrain classification, sound has been successfully integrated into robotic systems. For instance, bat-inspired echolocation has been used to accurately classify terrain types \cite{riopelle2018terrain}, such as grass, concrete, sand, and gravel, by using signal filtering and machine learning techniques. Similarly, acoustic sensors have been applied to real-world terrain classification tasks in robotics \cite{mason2024acoustic, christie2016acoustics}, and multimodal approaches combining sound data with cameras and foot sensors have improved semantic predictions \cite{xue2022sound, mason2024acoustic}. Despite its potential, sound has not been widely used to support traversability prediction. To address this gap, we incorporate acoustic vibrations into our mapping system to predict environmental conditions and the understanding of semantic information of terrains that are in contact with the robot's body.

%===============================================================================

%%%%%%%%%%%%%%%%%%%%%%%%%%%%%%%%%%%%%%%%%%%%%%%%%%%%%%%%%%%%%%%%%%%%%%%%%%%%%%%%
\section{Method}

\subsection{System Overview}
WildFusion aims to represent in-the-wild environments by constructing a multimodal 3D map. Our approach integrates multiple sensor groups as input and output modalities, including contact microphones, camera, LiDAR, tactile, and proprioception sensors. By exploring synergistic relationships between these modalities, our model gains a rich and accurate understanding of the complex environment. By combining these diverse perceptual signals, WildFusion addresses the limitations of traditional vision-based only methods and offers a more robust framework for scene representation and navigation.

\begin{figure}[t!]
    \centering\includegraphics[width=1\linewidth]{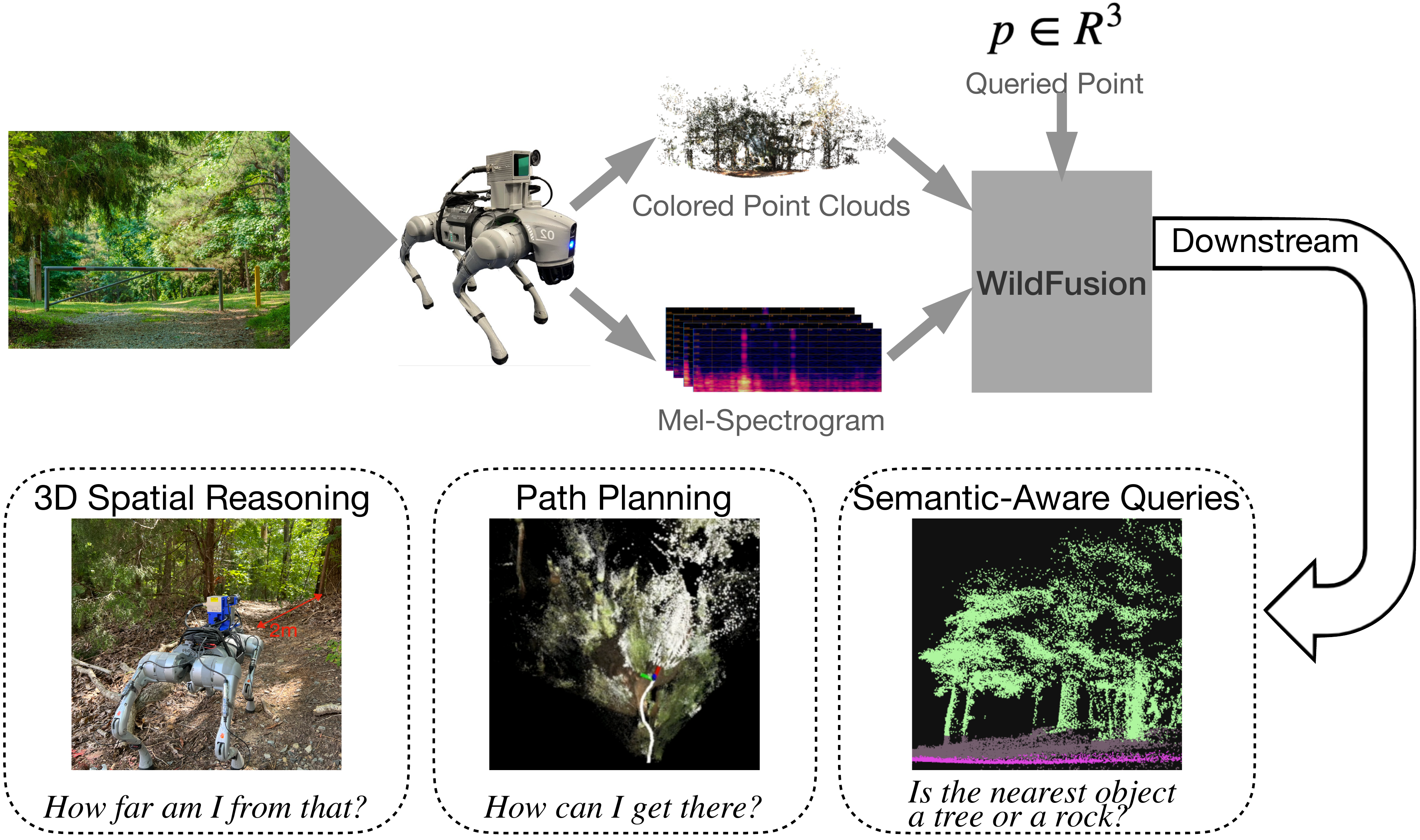}
    \caption{\textbf{System Overview:} WildFusion leverages signals from a RGB camera, LiDAR, contact microphones, tactile sensors, and an IMU to build rich multimodal scene representations. The learned representation enables the creation of dense maps with diverse features and guides legged robot navigation through continuous traversability scores.}
    \label{fig:overview}
\end{figure}

Our system (Fig.~\ref{fig:overview}) processes colored point clouds by fusing RGB images with LiDAR data, and Mel spectrograms from contact microphones, and random sample coordinates as query points. The system outputs semantic category, color, traversability, SDF, and confidence scores for each queried point. Intuitively, our system answers multiple environmental characteristics from multimodal perceptual inputs for the queried points. Therefore, WildFusion can be interpreted as a queried-based differentiable scene representation.
\vspace{-0.015cm}
\subsection{Integrated Multimodal Robotic Platform}
\begin{figure}[t!]
    \centering
    \includegraphics[width=0.5\linewidth]{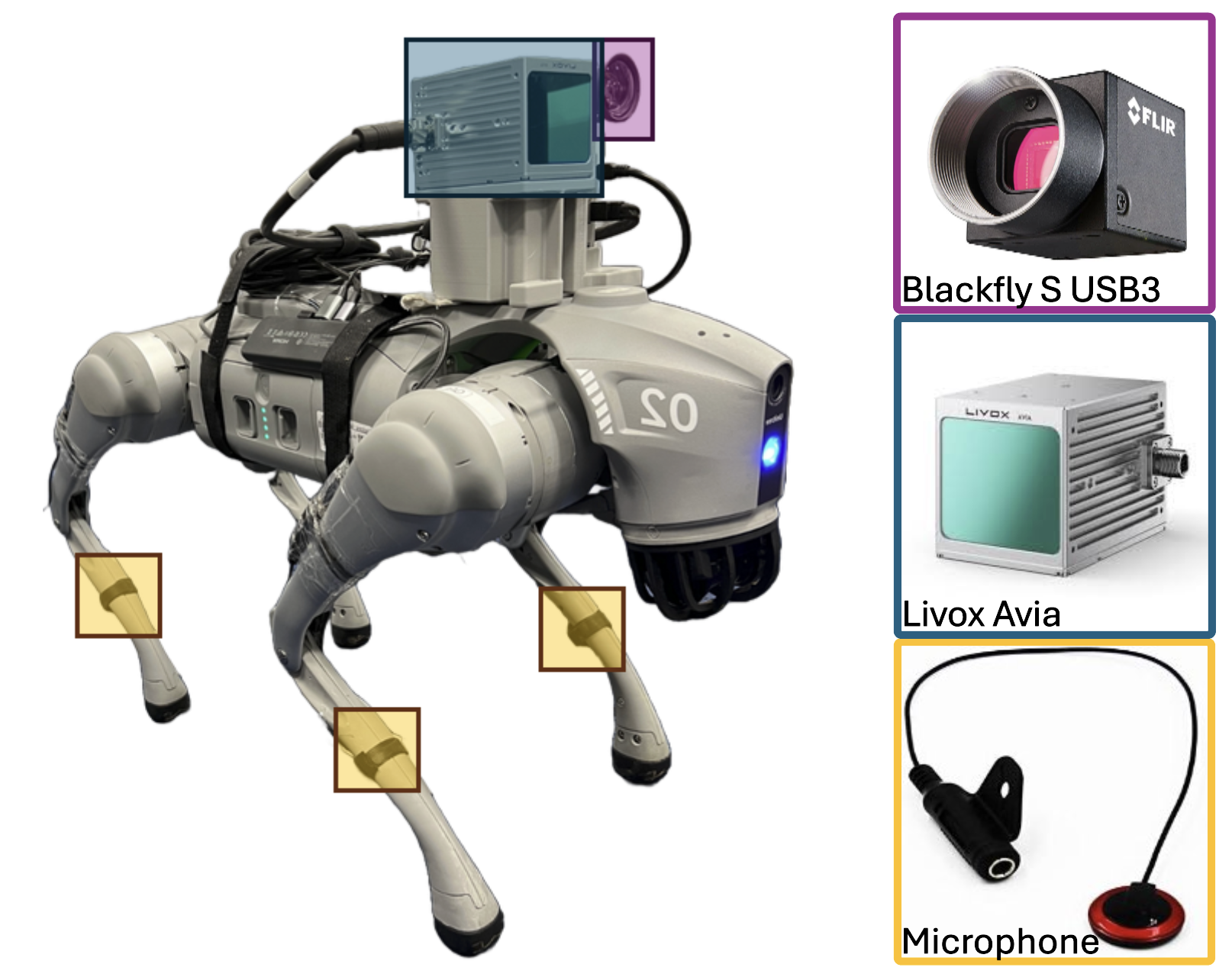}
    \caption{\textbf{Our Robotic Platform:} We integrate a monocular RGB camera, LiDAR and contact microphones on the Go2 platform. Proprioceptive data from  IMU and foot tactile sensors is also recorded to calculate ground truth for our model training.}
    \label{fig:go2Hardware}
\end{figure}

We built a legged robotic platform (Fig.~\ref{fig:go2Hardware}) capable of multimodal perception based on Unitree Go2. In addition to the original IMU and on-foot tactile sensors, we equipped the platform with several key modules. We first added a Blackfly S USB3 RGB camera due to its flexibility for lens mount, allowing us to optimize the field of view and depth of field with interchangeable lens. We paired it with a Tamron C-Mount 4 to 12\si{\milli\meter} Varifocal Manual Iris lens, which provides adjustable focal length and iris control for precise image capture tailored to our robot platform and forest setting. Additionally, we mounted a Livox AVIA solid-state LiDAR next to the camera with a 3D-printed frame. Our selected LiDAR sensor has a unique non-repetitive scanning pattern, which captures richer data over time compared to traditional mechanical LiDAR systems. This is especially useful in complex forest scenes, as it generates dense point clouds that improve the accuracy of ground-truth 3D information such as SDF. The sensor is also lightweight and cost-effective, offering higher field-of-view coverage within 0.3\si{\second} at just a fraction of the cost of traditional 64-line LiDARs.

As the robot moves across different terrains, contact between its feet and the ground will generate distinct vibrations. To capture these physical interactions without interference from environmental noises, we chose piezoelectric contact microphones and mounted them on the robot's calves. All sensors are connected to the onboard computer with synchronized timestamps. To ensure stable locomotion on challenging forest terrains, we trained a walking policy using a reinforcement learning strategy \cite{margolis2023walk}. The robot walks at a constant speed of 0.4\si[per-mode=symbol]{\meter\per\second} on a flat concrete floor, with variations in speed when traversing different terrain types.

\subsection{WildFusion Dataset}

The raw dataset includes RGB images, point clouds, audio recordings, foot contact forces, and IMU data collected from the quadruped robot over a $750\si{\meter}$ trajectory during a 30\si{\minute} walk through a forest inside Eno River State Park in NC, USA. This scene covers $90\%$ forested area with trees, leaves, tall vegetation, and fallen tree trunks, while the remaining $10\%$ consists of gravel or concrete paths. Due to the non-repeated scanning pattern of the AVIA LiDAR, each frame is generated by fusing data collected over a 2-second interval. In total, the processed dataset contains 551 frames, with 460 used for training, 50 for validation, and 41 for testing. Ground truth labels are provided for semantic segmentation, traversability, SDF, and confidence scores for each frame.

\textbf{Semnatic Labels} are manually assigned to points on or within objects, without differentiating between individual instances. \textbf{Color} is post-processed by converting it from RGB to LAB color space due to its uniformity of color perceptual distances and better handling of illumination variations. \textbf{Traversability} scores are calculated frame-by-frame using accelerometer and tactile sensor data. Accelerometer readings are normalized across all three axes, and their variance is used to assess movement instability. Tactile data is processed by normalizing the force distribution across four sensors and calculating its deviation from an ideal balanced state, which indicates stable ground contact from a previous calibration recording. The final traversability score for each frame is the product of the accelerometer variance and tactile deviation with a normalized continuous value between $0$ and $1$.

To compute the \textbf{SDF} from LiDAR data, points are uniformly sampled along each LiDAR ray, and the distance to the nearest surface point is calculated using a K-D Tree with 30 leaf nodes. Free space points are labeled based on their minimum distance to observed surface points, while negative distances are determined by sampling points beyond the termination of the LiDAR ray to account for the finite depth of objects. Confidence scores are derived from the SDF, where free space points are assigned a confidence score of 1. For points with negative SDF values, the confidence score decreases exponentially as the distance from the surface increases. Semantic labels and color data for free space points are set to NULL to reflect the absence of real-world information in these regions.

\subsection{WildFusion Model}

To overcome the limitations of traditional vision-based only methods for scene understanding in unstructured environments, WildFusion introduces a multimodal implicit scene representation learning framework. Our key idea is to not only condition multimodal signals from the input, but also ask the model to fuse these features to output multimodal environmental features. Furthermore, our model is established as implicit neural representations to enable dense predictions from sparse inputs by simply querying the model with coordinates from higher resolutions. The diverse data types complement one another, hence they enhance the map's details and utility for downstream tasks. For instance, traditional approaches that only predict SDF are limited by ambiguous object boundaries and insufficient information to learn complex geometries. Instead, we train our model to simultaneously predict color and semantics along with SDF, so that these additional modalities help inform clearer object boundaries. Confidence scores give our model the ability to assess and express its uncertainty about SDF predictions. Additionally, traversability predictions derived from acoustic vibrations can assist in predicting the geometry and semantics of the ground near the robot.

\begin{figure}[t!]
    \centering
    \includegraphics[width=1\linewidth]{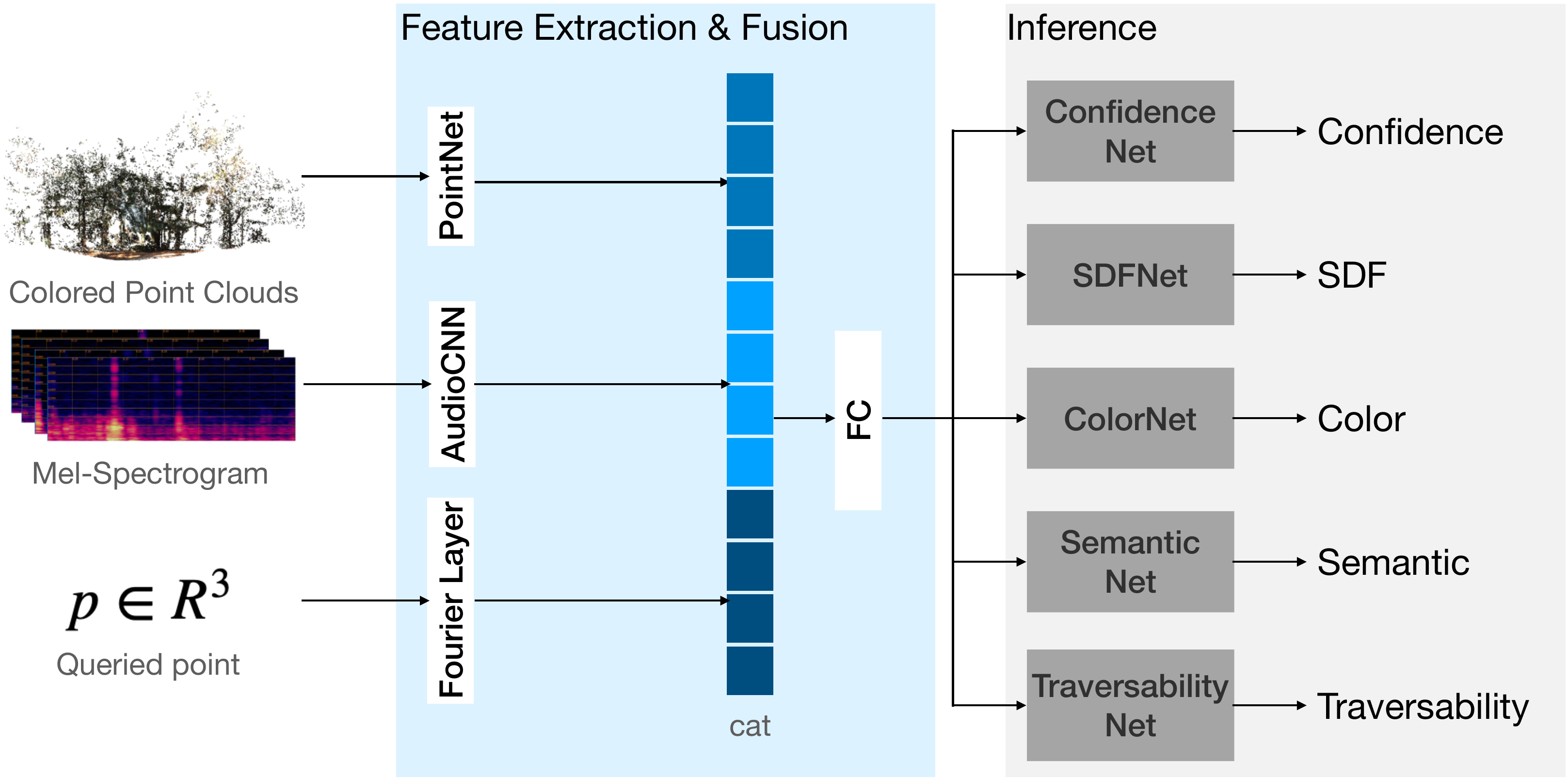}
    \caption{\textbf{WildFusion Model:} The model takes in a colored point cloud, Mel spectrogram, and queried points through a combination of encoders. The final outputs are derived through specialized sub-networks to predict confidence, SDF, color, semantics, and traversability.}
    \label{fig:architecture}
\end{figure}

We show our model architecture in Fig.~\ref{fig:architecture}. Specifically, the model takes colored point clouds, Mel-spectrograms, and queried coordinates as input. The queried point is passed through Fourier feature encoding \cite{tancik2020fourier} to efficiently capture high-frequency functions in low-dimensional data. Point clouds, consisting of 15,000 points, are processed using a modified PointNet \cite{qi2017pointnet}. We incorporate a T-Net for both input and feature transformations to ensure rotation invariance. We reduce the feature dimensions from 1024 to 512 and introduce residual connections to enhance gradient flow and convergence. The point cloud features capture both global and hierarchical information about the environment. Acoustic vibrations are converted to Mel-spectrograms, using a Fast Fourier Transform (FFT) window size of 2048 and 128 Mel bands, with a hop length of 512 and the highest frequency limited to 8,192Hz. These spectrograms, generated from 0.5-second audio segments, are then stacked along the leg channel and passed through several convolutional layers to extract relevant audio features.

Features extracted from different modalities are concatenated and fed into a comprehensive model layer, which predicts SDF, confidence score, color, traversability, and semantic through separate branches. The SDF and confidence score predictions utilize fully connected networks with residual skip connections to maintain smooth training dynamics. The SDF prediction applies a Tanh activation for smooth and bounded outputs, while the confidence score prediction uses a Sigmoid activation to map values between 0 and 1. For color and semantic label predictions, fully connected layers with ReLU are used to encourage efficient learning. Minimal dropout is applied in these branches to prevent overfitting while ensuring the model generalizes well.

We train our model using end-to-end supervised learning on our dataset. We minimize the following loss function which is a weighted sum  of the losses from each branch:

\begin{equation*}
\begin{aligned}
L = &\lambda_{\text{1}} L_{\text{SDF}} + \lambda_{\text{2}} L_{\text{eikonal}} 
 +\lambda_{\text{3}} L_{\text{confidence}} \\
& + \lambda_{\text{4}} L_{\text{semantics}} + \lambda_{\text{5}} L_{\text{color}} + \lambda_{\text{6}} L_{\text{traversability}} 
\end{aligned}
\end{equation*}

where \(\lambda\) values are constants that balance different losses. $L_{\text{SDF}}$ is Huber loss and we also include a small Eikonal loss \cite{sitzmann2020implicit} to ensure the gradient norm of SDF is close to one. This encourages the model to learn a geometrically consistent implicit surface representation. \(L_{\text{confidence}}\) helps focus on uncertain regions by assigning different weights based on whether the target confidence equals one:
\[
\frac{1}{N} \sum_{i=1}^{N} \left[ \alpha \cdot \mathbb{I}(c_i = 1) \cdot (\hat{c}_i - c_i)^2 + \beta \cdot \mathbb{I}(c_i \neq 1) \cdot (\hat{c}_i - c_i)^2 \right]
\] where \( \hat{c}_i \) is the predicted confidence for the $i$\textsuperscript{th} sample, \( c_i \) is the ground truth confidence for the $i$\textsuperscript{th} sample, \( \mathbb{I}(\cdot) \) is an indicator function, and \( \alpha \) and \( \beta \) are weights corresponding to whether the confidence is 1 or not.  \(L_{\text{semantics}}\) and  \(L_{\text{color}}\) are cross entropy loss, and the \(L_{\text{traversability}}\) uses mean squared error loss. Our model was trained on $8 \times$ NVIDIA A6000 GPUs for about 9 hours. We will open source all of our dataset, hardware integration solution, and code.

%%%%%%%%%%%%%%%%%%%%%%%%%%%%%%%%%%%%%%%%%%%%%%%%%%%%%%%%%%%%%%%%%%%%%%%%%%%%%%%%

\section{Experiment}

\subsection{Geometry, Confidence, Color, and Semantics}

\begin{table}[t!]
\caption{Quantitative Results}
\vspace{-0.3cm}
\label{tab:Quantitative}
\begin{center}
\resizebox{0.48\textwidth}{!}{%
\begin{tabular}{llcc}
\hline
                               &       & \begin{tabular}[c]{@{}c@{}}Seen Scenes \\ (w/ diff. viewpoints)\end{tabular} & Unseen Scenes \\ \hline
\multirow{3}{*}{Color}         & MSE   & 0.056                                                                        & 0.064         \\
                               & MAE   & 0.149                                                                        & 0.167         \\
                               & PSNR  & 13.690                                                                       & 12.258        \\ \hline
\multirow{2}{*}{Geometry}      & Haus. & 5.621                                                                        & 6.288         \\
                               & Cha.  & 0.072                                                                        & 0.079         \\ \hline
\multirow{5}{*}{Semantic}      & Acc.  & 0.886                                                                        & 0.755         \\
                               & Pre.  & 0.924                                                                        & 0.808         \\
                               & Re.   & 0.891                                                                        & 0.755         \\
                               & F1    & 0.907                                                                        & 0.726         \\
                               & IoU   & 0.825                                                                        & 0.692         \\ \hline
\multicolumn{1}{c}{confidence} & ECE   & 0.189                                                                        & 0.217         \\ \hline
\end{tabular}
}
\end{center}
\end{table}

\begin{figure*}[]
    \centering
    \includegraphics[width=0.71\linewidth]{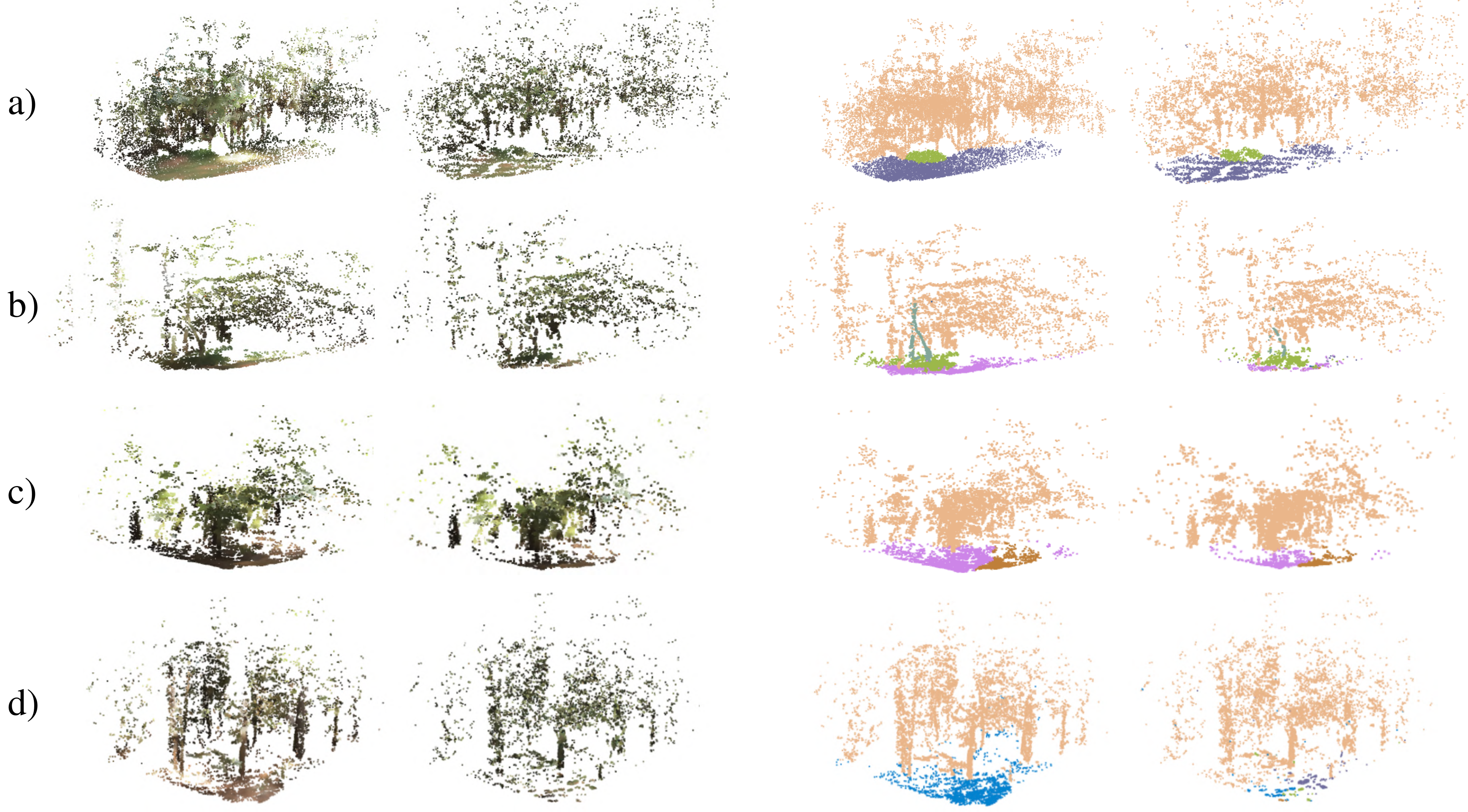}
    \caption{Visualization comparing geometry, color, and semantic predictions: The column 1 and 3 display the ground truth geometry, color, and semantics, while the column 2 and 4 show the corresponding predictions. a) and b) represent previously visited location from a different viewpoint. c) and d) represent unvisited location during training.}
    \label{fig:geoColorSem}
    \vspace{-0.5cm}
\end{figure*}
To evaluate the performance of our model, we conducted experiments on two groups of test sets: previously visited locations but observed from slightly different viewpoints and entirely unvisited locations. The previously visited locations with novel views assess the prediction consistency, while the unvisited locations test the model's generalizability and robustness. We sampled 30,000 points for each frame to get the prediction result. To ensure the validity of our results, we focused only on sample points with SDF values less than or equal to zero, as well as points with valid predicted semantic IDs. These criteria indicate that the points are either on the surface or inside objects, ensuring they provide meaningful data for geometry and semantic evaluation.

We employed a range of evaluation metrics across four key dimensions: color, geometry, confidence, and semantics. For color predictions, we report the mean-squared error (MSE), mean absolute error (MAE), and peak signal-to-noise ratio (PSNR). For geometry predictions, we calculated the Hausdorff distance to measure maximum point-based deviations and captured worst-case scenarios. We also computed the Chamfer distance to get average nearest-point distances. For semantic predictions, we applied standard classification metrics including accuracy, precision, recall, F1 score, and intersection over union (IoU). For confidence, we use expected calibration error (ECE).

As presented in Table~\ref{tab:Quantitative}, our model achieved high accuracy in predicting geometry, confidence, color, and semantics for previously visited locations with unseen viewpoints. For unseen scenes, while some discrepancies arose compared to the ground truth, the model still produced meaningful predictions. Fig.~\ref{fig:geoColorSem} visualizes these results, indicating that although performance slightly decreases in unseen environments, the model maintained a reasonable level of accuracy and successfully generalized to new scenes.

\subsection{Modality Studies}

\begin{table}[t!]
\caption{Modality Studies}
\vspace{-0.4cm}
\label{tab:ablationSDF}
\begin{center}
\resizebox{0.48\textwidth}{!}{%
\begin{tabular}{llccccc}
\hline
                          &       & \begin{tabular}[c]{@{}c@{}}Full \\ mode\end{tabular} & \begin{tabular}[c]{@{}c@{}}SDF,\\ Semantic,\\ Color\end{tabular} & \begin{tabular}[c]{@{}c@{}}SDF, \\ Conf. \\ Color\end{tabular} & \begin{tabular}[c]{@{}c@{}}SDF, \\ Conf.,\\ semantic\end{tabular} & \begin{tabular}[c]{@{}c@{}}SDF,\\ Conf.\end{tabular} \\ \hline
\multirow{3}{*}{Color}    & MSE   & 0.060                                                & 0.061                                                            & 0.066                                                          & -                                                                 & -                                                    \\
                          & MAE   & 0.162                                                & 0.172                                                            & 0.176                                                          & -                                                                 & -                                                    \\
                          & PSNR  & 12.633                                               & 12.518                                                           & 12.097                                                         & -                                                                 & -                                                    \\ \hline
\multirow{2}{*}{Geometry} & Haus. & 6.166                                                & 6.499                                                            & 6.512                                                          & 6.512                                                             & 7.514                                                \\
                          & Cha.  & 0.068                                                & 0.081                                                            & 0.082                                                          & 0.083                                                             & 0.083                                                \\ \hline
\multirow{5}{*}{Semantic} & Acc.  & 0.769                                                & 0.717                                                            & -                                                              & 0.705                                                             & -                                                    \\
                          & Prec. & 0.851                                                & 0.807                                                            & -                                                              & 0.774                                                             & -                                                    \\
                          & Re.   & 0.763                                                & 0.717                                                            & -                                                              & 0.705                                                             & -                                                    \\
                          & F1    & 0.770                                                & 0.694                                                            & -                                                              & 0.693                                                             & -                                                    \\
                          & IoU   & 0.704                                                & 0.638                                                            & -                                                              & 0.619                                                             & -                                                    \\ \hline
Confidence                & ECE   & 0.205                                                & -                                                                & 0.79                                                           & 0.228                                                             & 0.778                                                \\ \hline
\end{tabular}
}
\end{center}
\vspace{-0.4cm}
\end{table}

Our hypothesis is that all modalities help the model learn richer and more accurate 3D representations. We ablated our design by removing various modalities and compared the results with our original model (Table~\ref{tab:ablationSDF}). Our results show that all modalities contribute to the overall performance. The absence of confidence scores introduced additional uncertainty into predictions, leading to less accurate predictions. Similarly, the removal of color or semantics leads to a drop in shape reconstruction accuracy, as evidenced by increased Chamfer and Hausdorff distances.  These two modalities show an interdependence, as both are crucial for capturing fine-grained details. Semantics, in particular, helped the model become more confident in its predictions. The consistently better performance of our full model, compared to the ablated versions, confirms the complementary nature of each modality in producing richer and more accurate 3D representations.

\subsection{Motion Planning and Traversability Analysis}
\begin{figure*}[t!]
    \centering
    \includegraphics[width=0.71\linewidth]{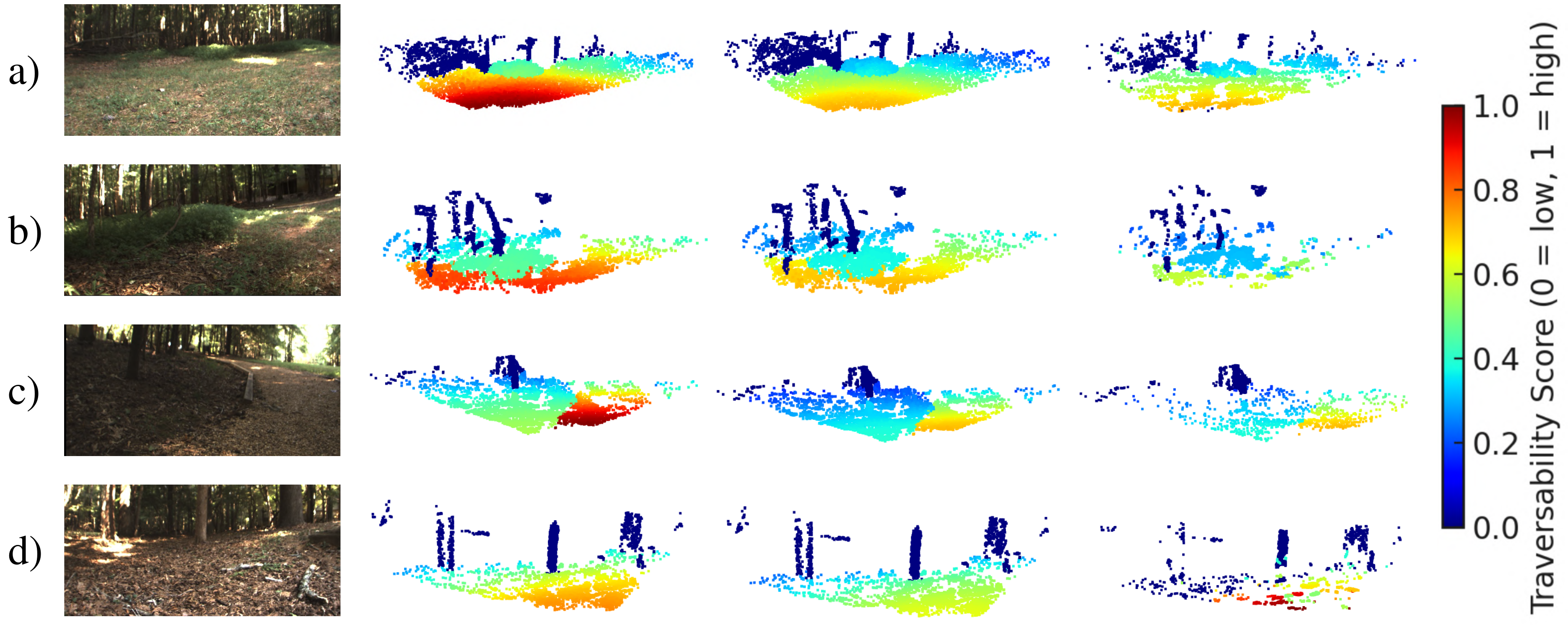}
    \caption{Traversability: Columns from left to right show the real scene, weight factor, ground truth traversability, and predicted traversability. The weight factor scales the single traversability score into point-wise values based on both semantic labels and distance from the robot. a) and b) represent previously visited location from a different viewpoint. c) and d) represent unvisited location during training. Blue indicates non-traversable areas, and red indicates fully traversable areas.}
    \label{fig:trav}
\vspace{-0.3cm}
\end{figure*}

Our model can be effectively used for downstream motion planning tasks, as we will demonstrate on our legged robot. Since the direct output from our model provides a single traversability score per scene, we refine this score to obtain pixel-level scores for motion planning by grounding it in scene context. First, we leverage the semantics of objects in the scene. Each semantic class in our dataset is manually assigned a traversability score, allowing us to generate a semantics-based traversability mask based on our predicted semantics. Second, because acoustics vibrations from foot-ground interactions influence our traversability prediction, we give higher weight to prediction near the robot's location. This results in a distance-based traversability mask. Specifically, this mask is created by taking the Hadamard product of the distance-based mask and a Gaussian distribution matrix centered around the robot. We empirically determined the variance to be six since the variance decides how quickly the weights decrease. The final pixel-level traversability scores are calculated as the Hadamard product of the semantic and distance-based masks, weighted by our model's prediction. 

\begin{figure}[t!]
    \centering
    \includegraphics[width=1\linewidth]{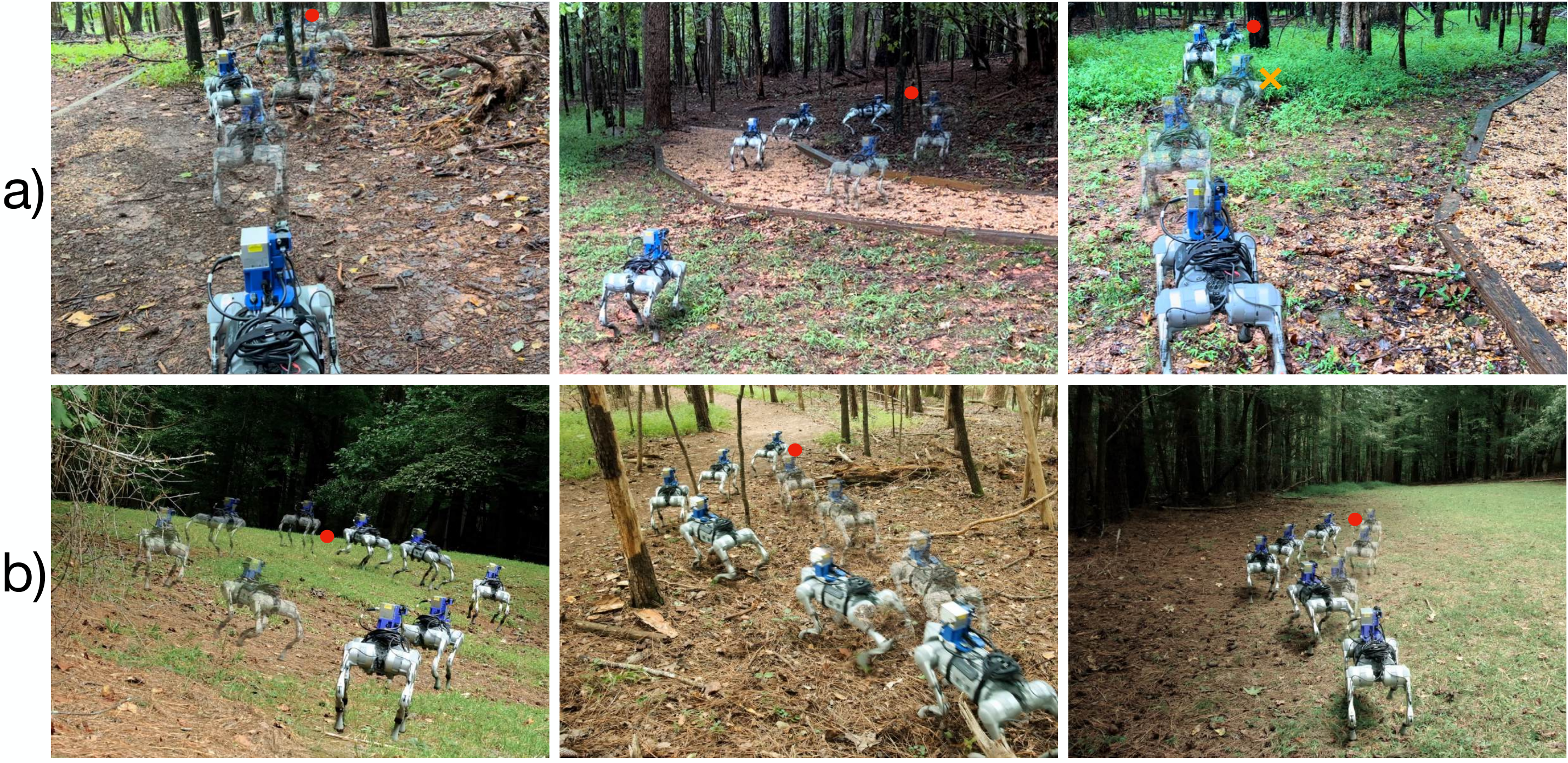}
    \caption{Real-world path planning: Compared the trajectory of our proposed model (solid robot) against baseline methods (60\% transparent robot). The red dot indicates the desired destination, and yellow cross mark means the robot stucked here. a) shows comparison against elevation-based costmap. b) shows comparison against semantic-based costmap.}
    \label{fig:realworld}
\end{figure}

Fig.~\ref{fig:trav} provides a visualization of weight factors and prediction results. The overall performance of traversability prediction aligned well with the ground truth, though specific values deviated by approximately $5\%$. This discrepancy is likely due to acoustic vibration data can only inform more accurate traversability of the near-feet grounds. One potential solution is to study longer dependencies from the acoustic data with more sophisticated network designs in the future.

We conducted a real-world experiment to evaluate how traversability prediction guides robot motion planning. We used an $A^*$ planner and projected our traversability scores as the costmap. Regions with lower traversability score has higher cost and vice versa. We compared our model with two baselines: an elevation map-based costmap and a variant of our proposed model that only predicts semantic information and assigns cost based on category.
Fig.~\ref{fig:realworld} shows the performance of our method against baselines in multiple scenarios. The elevation map based method showed the most limitations. It misidentified dense vegetation as obstacles and stopped moving forward. Moreover, it failed to differentiate between terrains with similar height, thus won't avoid the harder terrains. Our model performed well in both cases and would try to choose the safest route. Comparative analysis with the semantic-based method indicated our complete version of WildFusion enhanced the understanding of terrains by effectively guiding the robot toward flatter routes. In ambiguous scenarios, such as those combining solid ground with terrain littered with falling branches or grass, since the variation is different between them, our model's ability to discern distinct vibration patterns allowed better decision-making.
% On relatively simple and flat terrain, our model performs similarly to the baseline. However, in scenes with varied terrain types, our model consistently guides the robot to safer paths, such as gravel or concrete roads. In ambiguous situations, such as navigating through high vegetation, the baseline, which relies heavily on geometry, treats the vegetation as an obstacle, causing the robot to stop. In contrast, our model allows the robot to continue moving forward.

%%%%%%%%%%%%%%%%%%%%%%%%%%%%%%%%%%%%%%%%%%%%%%%%%%%%%%%%%%%%%%%%%%%%%%%%%%%%%%%%
\section{Conclusions, Limitations, and Future Work}

WildFusion presents a novel multimodal framework for implicit 3D reconstruction and navigation in unstructured outdoor environments. By integrating data from LiDAR, RGB camera, contact microphone, and proprioceptive sensors, WildFusion creates a rich pixel-level representation of complex terrains, which enhances environmental understanding and provides support for more efficient and reliable path planning. Our experiments show that this approach significantly improves navigation in a forest setting. 

For future work, expanding sensor modalities, such as incorporating humidity and thermal sensors, could provide a more holistic environmental understanding. Furthermore, refining our traversability prediction with more advanced models could lead to even greater accuracy by considering more dynamic information from the robot motions. Moreover, our current model requires offline computational resources. A future direction can study how to enable on-board training with edge computing techniques.


%%%%%%%%%%%%%%%%%%%%%%%%%%%%%%%%%%%%%%%%%%%%%%%%%%%%%%%%%%%%%%%%%%%%%%%%%%%%%%%%

%%%%%%%%%%%%%%%%%%%%%%%%%%%%%%%%%%%%%%%%%%%%%%%%%%%%%%%%%%%%%%%%%%%%%%%%%%%%%%%%



% \section*{ACKNOWLEDGMENT}

% The preferred spelling of the word ÒacknowledgmentÓ in America is without an ÒeÓ after the ÒgÓ. Avoid the stilted expression, ÒOne of us (R. B. G.) thanks . . .Ó  Instead, try ÒR. B. G. thanksÓ. Put sponsor acknowledgments in the unnumbered footnote on the first page.

%%%%%%%%%%%%%%%%%%%%%%%%%%%%%%%%%%%%%%%%%%%%%%%%%%%%%%%%%%%%%%%%%%%%%%%%%%%%%%%%

\bibliographystyle{IEEEtran}
\bibliography{wildfusion}

\end{document}